\documentclass[11pt, oneside]{article}   	
\usepackage[margin=.8in]{geometry}               		
\usepackage{graphicx}				
\usepackage{amssymb}
\usepackage{color}
\usepackage{amsmath}
\usepackage{appendix}
\usepackage{enumerate}
\usepackage{bbm}
\usepackage{mathabx}
\usepackage{hyperref}
\usepackage{amsthm}
\usepackage{authblk}
\usepackage[numbers]{natbib}

\usepackage{tikz}

\usetikzlibrary{shadows}
\tikzset{
diagonal fill/.style 2 args={fill=#2, path picture={
\fill[#1, sharp corners] (path picture bounding box.north west) -|
                         (path picture bounding box.south east) -- cycle;}},
reversed diagonal fill/.style 2 args={fill=#2, path picture={
\fill[#1, sharp corners] (path picture bounding box.south west) |- 
                         (path picture bounding box.north east) -- cycle;}}
}

\usetikzlibrary{arrows.meta}

\title{Capacity allocation through neural network layers}

\author{Jonathan Donier\footnote{jdonier@spotify.com}}
\affil{Spotify}

\begin{document}
\maketitle

\begin{abstract}
Capacity analysis has been recently introduced as a way to analyze how linear models distribute their modelling capacity across the input space. In this paper, we extend the notion of capacity allocation to the case of neural networks with non-linear layers. We show that under some hypotheses the problem is equivalent to linear capacity allocation, within some extended input space that factors in the non-linearities. We introduce the notion of layer decoupling, which quantifies the degree to which a non-linear activation decouples its outputs, and show that it plays a central role in capacity allocation through layers. In the highly non-linear limit where decoupling is total, we show that the propagation of capacity throughout the layers follows a simple markovian rule, which turns into a diffusion PDE in the limit of deep networks with residual layers. This allows us to recover some known results about deep neural networks, such as the size of the effective receptive field, or why ResNets avoid the shattering problem.
\end{abstract}

\section{Introduction}

Capacity allocation analysis was recently introduced in \cite{donier2018capacity} as a way to determine quantitatively which dependencies between its inputs and output(s) a trained model has focussed its modelling capacity on. The theory was developed in the context of linear models and $L^2$ loss, which corresponds to the Gaussian process prediction task. It was shown that the total model capacity corresponds to the number of independent parameters in the model, and that the capacity allocation could be broken down along any partition of the input space. One partition of particular interest is the spatial partition, i.e. the partition along the canonical dimensions of the input space. This allowed to perform spatial capacity allocation analysis, i.e. to determine which inputs the model has focussed on, and to which degree. This was then used to study two types of 1-dimensional architectures that are commonly used for audio modelling -- hierarchical models as in Wavenets \cite{van2016wavenet}, and recurrent models \cite{mehri2016samplernn, kalchbrenner2018efficient}.

The intuition behind such analysis is to provide tools to peek into the neural black boxes, and understand \textit{a posteriori} what a given architecture has tried to model, when trained on a given task. Rather than analyzing the network \emph{activations} as it is often done \cite{simonyan2013deep, zeiler2014visualizing, sellam2018deepbase, zintgraf2017visualizing}, the idea is to analyze \emph{parameters}. The extensive quantity thus defined, called the \emph{capacity} $\kappa$, turns out to have interesting properties. The idea vindicated in \cite{donier2018capacity} is that studying this quantity can lead to new insights on (deep) neural networks, as it is complementary to the value of the loss function: for example, it allows to pinpoint which dependencies the model has tried to capture in order to reduce the value of the loss. When the final goal is more complex than merely minimizing a scalar value, e.g. for content generation tasks, exercising some control over the model's inductive bias could turn out to be highly valuable, in particular when training full models is long and difficult.

One hypothesis that was put forward in \cite{donier2018capacity} is that the spatial dimension of architecture design (i.e. whether to use recurrent, convolutional, hierarchical models etc.) can be decoupled from its expressivity \cite{raghu2016expressive, poole2016exponential, guss2018characterizing} (i.e. the complexity of the dependencies it can model), and accordingly the analysis was focussed on the spatial aspects in the linear case. Still, when the model takes a linear form in some \emph{feature space} (obtained e.g. by some non-linear transformation of the inputs), capacity allocation was shown to make sense in that feature space. However, it was suggested that propagating this capacity allocation to the input space is non-trivial in general.

In this paper, we take capacity allocation analysis one step further and investigate the other dimension of the problem by propagating capacity allocation through non-linearities \emph{into the input space}. More specifically, we analyze the case of neural network layers, which transform their inputs $Y$ according to:

\begin{equation}
\phi(Y) = f(P^TY).
\end{equation}

\noindent We approach the problem by reducing it to a linear case, by factoring out the non-linearity into some augmented input space. While this requires exponentially large augmented input spaces, we show that two limiting cases are tractable, namely:

\begin{enumerate}[(i)]
\item for \emph{linear} layers (i.e. when $f$ is the identity function), where one recovers standard capacity allocation in the input space,
\item for \emph{pseudo-random} layers (in a sense defined below), where the capacity in the input space can be derived explicitly from the capacity in the feature space.
\end{enumerate}

\noindent If we introduce the notion of layer \emph{decoupling} to quantify how much a non-linear activation decouples its outputs, then these two types of layers are two extreme cases of decoupling: linear activations do not modify their inputs and hence perform no decoupling, while pseudo-random activations achieve a full decoupling. On this scale, usual non-linear activations lie somewhere in between, and the extent to which they decouple their outputs is a key element that defines how capacity is transferred from feature space to input space.

The end of this paper focusses on the pseudo-random end of the spectrum, where non-linearity plays a full role (the linear case was already discussed in \cite{donier2018capacity}). We use this framework to analyze deep networks that are made of a large number of such layers, and show that capacity propagation through layers can be described by some linear \emph{capacity network}, which acts backwards from the output space to the input space, and which can be approximated by a simple diffusion PDE in some scaling limits. This allows us to use the power of PDE analysis to study deep neural networks, and recover in passing some well-known results about the effective receptive field and the shattering problem.

We start by introducing the capacity allocation problem in the case of single neural network layers in Section \ref{sec:single_layer}, where we introduce the notions of augmented input space and layer decoupling. We then study two limiting cases: the linear case and the pseudo-random case, where the capacity in the input space can be computed directly from the capacity in the feature space under some independence hypotheses. We then build upon these results to turn to the case of deep networks in Section \ref{sec:multiple_layer}, where we show that in some limit capacity propagation through layers follows a diffusion PDE. We finally leverage these findings in Sections \ref{sec:applications} where we address some fundamental questions about the behaviour of deep networks.

\section{Capacity allocation of single neural network layers}\label{sec:single_layer}

\subsection{The problem}

The concept of capacity in a feature space was introduced in \cite{donier2018capacity} and defined as follows. For some function $\phi$ that maps some input $Y\in\mathbb{R}^n$ to some intermediate space $\mathbb{R}^m$, we consider models of the form $\varphi_W(Y) = A_W^T\phi(Y)$, where the vector of coefficients $A_W$ is parametrized by $W\in \mathbb{R}^p$. The coefficients space is defined as  $\mathcal{A} = \left\{A_W\in\mathbb{R}^m \mid W\in \mathbb{R}^p \right\}$. Since the model is linear in the space of $\phi(Y)$ (a.k.a. the \emph{feature space}), capacity allocation in that space can be computed by applying linear capacity analysis. The capacity allocated to some subspace $\mathcal{S}^\phi$ of the feature space is noted $\kappa^\phi(\mathcal{S}^\phi)$. The next step is then to compute the capacity allocation in the \emph{input space}, but this question remains unanswered in general. Here we focus on a special class of mapping functions $\phi$, namely single neural network layers that can be written as:

\begin{equation}
\phi(Y) = f(P^TY),
\end{equation}

\noindent where $P = \{p_1, ..., p_m \}\in \mathbb{R}^{n\times m}$ is a linear transformation such that the vectors $p_i$ are all distinct, and $f(\cdot)$ is a pointwise non-linear function (such as ReLU, sigmoid, etc.). We assume for simplicity that the $p_i$'s are also normalized to $\|p_i \|_2 = 1$. The full model therefore reads:

\begin{equation}\label{eq:model}
\varphi_W(Y) = A_W^T f(P^TY),
\end{equation}

We will be mostly interested in how $A_W$ allocated its capacity in the input space, assuming that $P$ is constant (which we assume for simplicity, but happens to be a common practice that can give good results \cite{rosenfeld2018intriguing, saxe2011random}). We will briefly consider the case where $P$ is itself learned in Appendix \ref{sec:P}, and suggest that it might be handled in a similar way -- at least in some circumstances.


\subsection{Rewriting the problem in the augmented input space}

The main idea is to rewrite the non-linear function $f(\cdot)$ from Eq. (\ref{eq:model}) as:

\begin{equation}
f(z) = \eta_z z,
\end{equation}

\noindent where $\eta_z$ is by definition equal to $f(z) / z$. We can then rewrite $\varphi$ as:

\begin{equation}\label{eq:augmented_model}
\begin{aligned}
\varphi(Y) &=A^T\left( \eta \odot P^TY\right), \quad \eta = (\eta_1, ..., \eta_m).\\
:&= A^T\tilde{P}^T\tilde{Y}
\end{aligned}
\end{equation}

\noindent where $\eta_i$ is a shortcut for $\eta_{(P^TY)_i}$ and:

\begin{equation}\label{eq:augmented_model_detail}
\tilde{P} = 
\begin{bmatrix}
&p_1 &0 &\hdots &0\\
&0 &p_2& \hdots &0\\
&\vdots&\vdots&\ddots&\vdots\\
&0 &0& \hdots &p_m
  \end{bmatrix} \in \mathbb{R}^{nm \times m}, \quad 
  \tilde{Y} = 
\begin{bmatrix}
\eta_1 y_1\\
 \vdots\\
  \eta_1y_n\\
   \vdots\\
   \eta_my_1\\  
    \vdots\\
   \eta_my_n\\  
\end{bmatrix}\in \mathbb{R}^{nm}.
\end{equation}

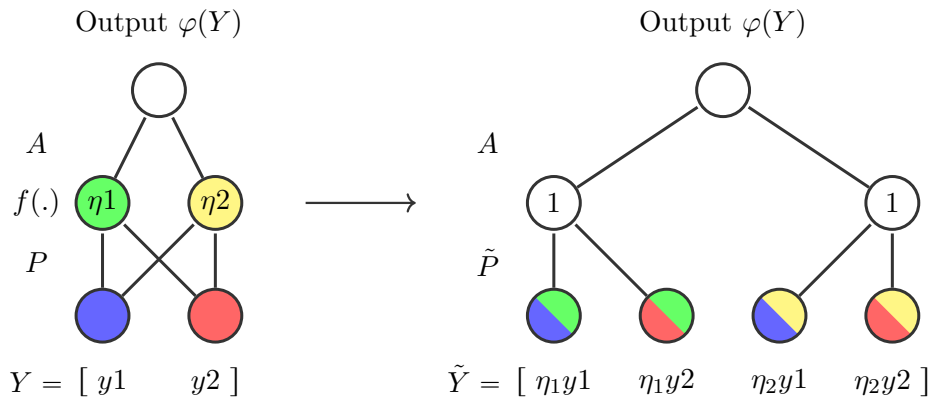
\begin{figure}[!b]

\def\layersep{1.5}
\def\nodesep{1.5}
\def\nodesize{20pt}
\def\margins{25}

\begin{tikzpicture}[shorten >=1pt,->,draw=black!80, node distance=\layersep, line width=0.4mm]
    \tikzstyle{every pin edge}=[<-,shorten <=1pt]
    \tikzstyle{neuron}=[circle,draw=black!80,minimum size=\nodesize,inner sep=0pt,line width=0.4mm]
    \tikzstyle{input neuron}=[neuron,fill=yellow!50];
    \tikzstyle{invisible neuron}=[neuron,minimum size=0pt];
    \tikzstyle{output neuron}=[neuron,fill=red!60];
    \tikzstyle{hidden neuron}=[neuron];
    \tikzstyle{annot} = [ text centered]

        \node[neuron, fill=blue!60] (n1) at (0*\nodesep,0) {};
        \node[neuron, fill=red!60] (n2) at (1*\nodesep,0) {};
        
        \node[neuron, fill=green!60] (n3) at (0*\nodesep,\layersep) {$\eta1$};
        \node[neuron, fill=yellow!60] (n4) at (1*\nodesep,\layersep) {$\eta2$};
        
        \node[neuron] (n5) at (0.5*\nodesep,2*\layersep) {};

        \node[neuron, diagonal fill={green!60}{blue!60}] (m1) at (4*\nodesep,0) {};
        \node[neuron, diagonal fill={green!60}{red!60}] (m2) at (5*\nodesep,0) {};
        \node[neuron, diagonal fill={yellow!60}{blue!60}] (m3) at (6*\nodesep,0) {};
        \node[neuron, diagonal fill={yellow!60}{red!60}] (m4) at (7*\nodesep,0) {};
        
        \node[neuron] (m5) at (4*\nodesep,\layersep) {1};
        \node[neuron] (m6) at (7*\nodesep,\layersep) {1};
        
        \node[neuron] (m7) at (5.5*\nodesep,2*\layersep) {};

   \path[-] (n1) edge (n3);
   \path[-] (n1) edge (n4);
   \path[-] (n2) edge (n3);
   \path[-] (n2) edge (n4);
   \path[-] (n3) edge (n5);
   \path[-] (n4) edge (n5);

   \path[-] (m1) edge (m5);
   \path[-] (m2) edge (m5);
   \path[-] (m3) edge (m6);
   \path[-] (m4) edge (m6);
   \path[-] (m5) edge (m7);
   \path[-] (m6) edge (m7);

    \node[annot,below of=n1, node distance=\margins] (y1) {$[~y1$};
   \node[annot,below of=n2, node distance=\margins] (y2) {$y2~]$};
   
   \node[annot,left of=y1, node distance=\margins] (y) {$Y=$};
   
   \node[annot,left of=n3, node distance=\margins] (f) {$f(.)$};
   \node[annot,below of=f, node distance=0.9*\margins] (P) {$P$};
    \node[annot,above of=f, node distance=0.9*\margins] (A){$A$};
    \node[annot,above of=n5, node distance=\margins] {Output $\varphi(Y)$};
    
    \node[annot,below of=m1, node distance=\margins] (ny1) {$[~\eta_1 y1$};
   \node[annot,below of=m2, node distance=\margins] (ny2) {$\eta_1 y2$};
    \node[annot,below of=m3, node distance=\margins] (ny3) {$\eta_2 y1$};
   \node[annot,below of=m4, node distance=\margins] (ny4) {$\eta_2 y2~]$};
   
    \node[annot,below of=m1, node distance=0.95*\margins] (fakeny1) {};
   \node[annot,left of=fakeny1, node distance=1.2*\margins] (yt) {$\tilde{Y}=$};
   \node[annot,left of=m5, node distance=\margins] (empty2) {};

   \node[annot,below of=empty2, node distance=0.9*\margins] (P) {$\tilde{P}$};
    \node[annot,above of=empty2, node distance=0.9*\margins] (A){$A$};
    \node[annot,above of=m7, node distance=\margins] {Output $\varphi(Y)$};
    
   \path[line width=0.3mm] (1.8*\nodesep, \layersep) edge (2.8*\nodesep, \layersep) ;
   
\end{tikzpicture}
\centering
\caption{\label{fig:decoupling} Linearizing a layer by decoupling the paths. The non-linearity $\eta$ is factored in the augmented input space, which allows us to re-write $A^T f(P^TY)$ as $A^T\tilde{P}^T\tilde{Y}$.}

\end{figure}

\noindent What we have done here is simply to decouple the components of $Y$ according to the particular $\eta_i$ that multiplies them when going through $f$ (see Figure \ref{fig:decoupling} for an illustration). 

Now let us consider the optimal model that satisfies:

\begin{equation}\label{eq:surrogate}
\varphi^*(Y) = \underset{\varphi}{\text{argmin }} \mathbb{E}\left[ \left(\varphi(Y) -\varphi_W(Y)\right)^2 \right] := \mathcal{L}_W
\end{equation}

\noindent which can by definition be written in the form $\varphi^*(Y) = A^{*T}  \tilde{P}^{*T} \tilde{Y}$. The idea is to use $\varphi^*$ as a surrogate ``true'' process to replace $\varphi_W$ in the calculation of the capacity.\footnote{Note from \cite{donier2018capacity} that given some optimal model, the capacity allocation does not depend on the true model. We can therefore use a surrogate of the true model for the purpose of computing the capacity allocation, e.g. the optimal model itself, which by definition minimizes both losses in Eq. (\ref{eq:surrogate}) and Eq. (\ref{eq:loss_tilde}).} The $L^2$ loss function associated to this process can then be written as:

\begin{equation}\label{eq:loss_tilde}
\begin{aligned}
\mathcal{L} &= \mathbb{E}\left[ \left(\varphi(Y) -\varphi^*(Y)\right)^2 \right] \\
&= \left(\tilde{P}A - \tilde{P}^*A^*\right) ^T \tilde{\Sigma}  \left(\tilde{P}A - \tilde{P}^*A^*\right)
\end{aligned}
\end{equation}

\noindent where $\tilde{\Sigma} := \mathbb{E}( \tilde{Y}\tilde{Y}^T )$ is the covariance matrix in the augmented input space. Cancelling the derivative of $\mathcal{L}$ with respect to the model parameters gives:

\begin{equation}\label{eq:capacity_augmented}
\begin{aligned}
\frac{\partial \mathcal{L}}{\partial W}  = 0 \quad 
& \Leftrightarrow \quad \tilde{K}^T\tilde{X} = 0, \quad \quad \tilde{K} :=   \tilde{\Sigma}  \frac{\partial \tilde{P}A}{\partial W} \in \mathbb{R}^{nm\times p}
\end{aligned}
\end{equation}

\noindent where $\tilde{K}$ is the capacity matrix in the augmented input space and where we have defined $\tilde{X} := \left(\tilde{P}A - \tilde{P}^*A^*\right)$ as the error in the space of the coefficients applied to $\tilde{Y}$. Note that the surrogate process $\varphi^*$ does not intervene in the expression of $\tilde{K}$, as expected. We have recovered the exact same equations as in the linear case \cite{donier2018capacity}, but applied in the \emph{augmented input space} (i.e. the space of $\tilde{Y}$).

Note that the above equivalence holds since $\tilde{\Sigma}$ does not depend on the model parameters, which follows from the fact that $\eta$ does not depend on $A$. If $P$ is itself learned, computing its capacity allocated in the input space is trickier as $\eta$ and therefore $\tilde{\Sigma}$ depend on $P$. We however suggest in Appendix \ref{sec:P} that the above equation might hold in general, or at least in some cases (including the linear case but also piecewise linear functions like ReLU, leaky ReLU or absolute value).

\subsection{Layer decoupling} \label{sec:decoupling}

The intuition that underlies the above reformulation of the problem in the augmented input space is that a pointwise (non-linear) function $f$ \emph{decouples} its outputs from each other. The degree of decoupling however depends on the particular activation $f$ that is used. Below we consider several types of activations and analyze their effects on $\tilde{\Sigma}$.

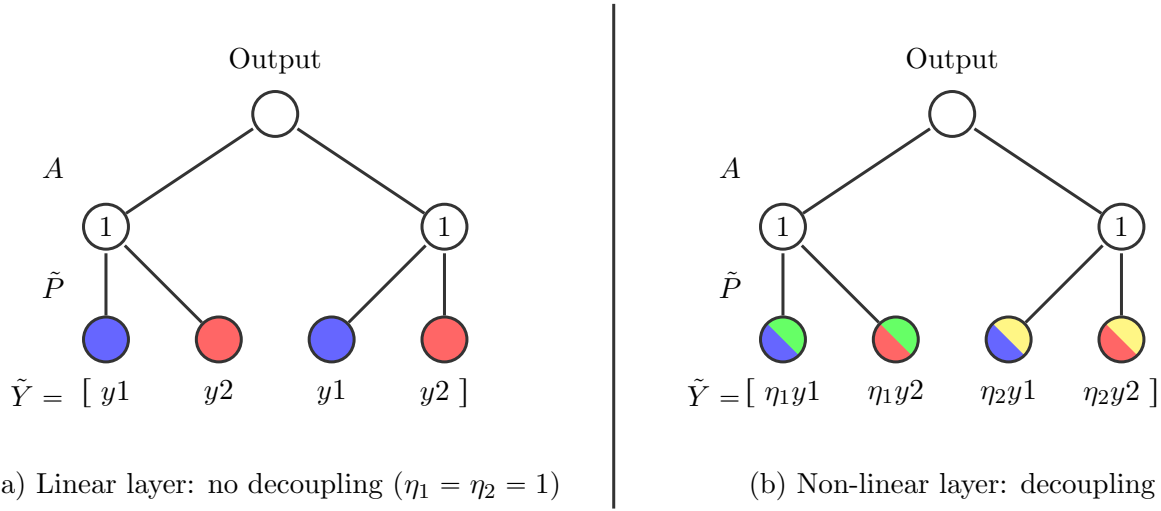
\begin{figure}[!t]

\def\layersep{1.5}
\def\nodesep{1.5}
\def\nodesize{17pt}
\def\margins{20}

\begin{tikzpicture}[shorten >=1pt,->,draw=black!80, node distance=\layersep,line width=0.4mm]
    \tikzstyle{every pin edge}=[<-,shorten <=1pt]
    \tikzstyle{neuron}=[circle,draw=black!80,minimum size=\nodesize,inner sep=0pt,line width=0.4mm]
    \tikzstyle{input neuron}=[neuron,fill=yellow!50];
    \tikzstyle{invisible neuron}=[neuron,minimum size=0pt];
    \tikzstyle{output neuron}=[neuron,fill=red!60];
    \tikzstyle{hidden neuron}=[neuron];
    \tikzstyle{annot} = [ text centered]

        \node[neuron, fill=blue!60] (n1) at (0*\nodesep,0) {};
        \node[neuron, fill=red!60] (n2) at (1*\nodesep,0) {};
        \node[neuron, fill=blue!60] (n3) at (2*\nodesep,0) {};
        \node[neuron, fill=red!60] (n4) at (3*\nodesep,0) {};
        
        \node[neuron] (n5) at (0*\nodesep,\layersep) {1};
        \node[neuron] (n6) at (3*\nodesep,\layersep) {1};
        
        \node[neuron] (n7) at (1.5*\nodesep,2*\layersep) {};

        \node[neuron, diagonal fill={green!60}{blue!60}] (m1) at (6*\nodesep,0) {};
        \node[neuron, diagonal fill={green!60}{red!60}] (m2) at (7*\nodesep,0) {};
        \node[neuron, diagonal fill={yellow!60}{blue!60}] (m3) at (8*\nodesep,0) {};
        \node[neuron, diagonal fill={yellow!60}{red!60}] (m4) at (9*\nodesep,0) {};
        
        \node[neuron] (m5) at (6*\nodesep,\layersep) {1};
        \node[neuron] (m6) at (9*\nodesep,\layersep) {1};
        
        \node[neuron] (m7) at (7.5*\nodesep,2*\layersep) {};


   \path[-] (n1) edge (n5);
   \path[-] (n2) edge (n5);
   \path[-] (n3) edge (n6);
   \path[-] (n4) edge (n6);
   \path[-] (n5) edge (n7);
   \path[-] (n6) edge (n7);

   \path[-] (m1) edge (m5);
   \path[-] (m2) edge (m5);
   \path[-] (m3) edge (m6);
   \path[-] (m4) edge (m6);
   \path[-] (m5) edge (m7);
   \path[-] (m6) edge (m7);

    \node[annot,below of=n1, node distance=\margins] (ny1) {$[~ y1$};
   \node[annot,below of=n2, node distance=\margins] (ny2) {$y2$};
    \node[annot,below of=n3, node distance=\margins] (ny3) {$y1$};
   \node[annot,below of=n4, node distance=\margins] (ny4) {$y2~]$};
   \node[annot,left of=ny1, node distance=1.3*\margins] (nyt) {$\tilde{Y}=$};
   \node[annot,left of=n5, node distance=\margins] (nempty) {};

   \node[annot,below of=nempty, node distance=1.1*\margins] (nP) {$\tilde{P}$};
    \node[annot,above of=nempty, node distance=1.1*\margins] (nA){$A$};
    \node[annot,above of=n7, node distance=\margins] {Output};
    
    \node[annot,below of=m1, node distance=\margins] (my1) {$[~\eta_1 y1$};
   \node[annot,below of=m2, node distance=\margins] (my2) {$\eta_1 y2$};
    \node[annot,below of=m3, node distance=\margins] (my3) {$\eta_2 y1$};
   \node[annot,below of=m4, node distance=\margins] (my4) {$\eta_2 y2~]$};
   \node[annot,left of=my1, node distance=1.3*\margins] (myt) {$\tilde{Y}=$};
   \node[annot,left of=m5, node distance=\margins] (mempty) {};

   \node[annot,below of=mempty, node distance=1.1*\margins] (mP) {$\tilde{P}$};
    \node[annot,above of=mempty, node distance=1.1*\margins] (mA){$A$};
    \node[annot,above of=m7, node distance=\margins] {Output};
    
    \path[-] (4.5*\nodesep, -1.5*\layersep) edge (4.5*\nodesep, 3*\layersep);
    
        \node[annot] at (1.5*\nodesep,-1.3*\layersep) {(a) Linear layer: no decoupling ($\eta_1=\eta_2=1$)};
        \node[annot] at (7.5*\nodesep,-1.3*\layersep) {(b) Non-linear layer: decoupling};

\end{tikzpicture}

\centering
\caption{\label{fig:linear_vs_random} \textit{(left)} For linear layers where $\eta_1=\eta_2=1$, no decoupling happens and the inputs are merely repeated. In this degenerate case, the capacity in the augmented input space matches the capacity in the original input space.  \textit{(right)} For non-linear layers, the inputs are decoupled by multiplication with the $\eta$'s. In the pseudo-random limiting case, the left (green) and the right (yellow) blocks become independent.}

\end{figure}

\paragraph{Linear activations} Choosing $f$ as the identity leads to no decoupling at all:

\begin{equation}\label{eq:sigma_plain}
\forall y, ~\eta_y = 1 \quad \Rightarrow \quad  \tilde{\Sigma} = 
\begin{bmatrix}
&\Sigma &\Sigma &\hdots &\Sigma\\
&\Sigma &\Sigma& \hdots &\Sigma\\
&\vdots&\vdots&\ddots&\vdots\\
&\Sigma &\Sigma& \hdots &\Sigma
  \end{bmatrix} \in \mathbb{R}^{nm\times nm}.
  \end{equation}
  
 \noindent i.e. the original inputs are simply repeated identically $m$ times. The correlation matrix is thus repeated $m$ times in both directions (see Figure \ref{fig:linear_vs_random}, left).

\paragraph{Pseudo-random activations}  The other extreme would be an activation function whose outputs are fully decoupled (which we call \emph{pseudo-random} activations):

\begin{equation}\label{eq:chaotic_sigma}
\mathbb{E}_z(\eta_z)=0, \quad \mathbb{E}_{z_1, z_2}(\eta_{z_1}\eta_{z_2}) = \sigma^2 \delta_{z_1-z_2} \quad \Rightarrow \quad \tilde{\Sigma} = \sigma^2
\begin{bmatrix}
&\Sigma &0 &\hdots &0\\
&0 &\Sigma& \hdots &0\\
&\vdots&\vdots&\ddots&\vdots\\
&0 &0& \hdots &\Sigma
  \end{bmatrix} \in\mathbb{R}^{nm\times nm},
  \end{equation}
  
\noindent where we will often set $\sigma=1$ for simplicity. We use the terminology ``pseudo-random activation'' since it corresponds to a non-linear function $f$ that is nowhere continuous, and such that the outputs corresponding to two arbitrarily close inputs can be effectively considered to be independent.\footnote{Note that the process $\eta_z$ is only drawn \emph{once}. Therefore, if the same input is fed twice to the network, it will produce the same output. However, because $\eta_z$ is nowhere continuous, two slightly different inputs might produce very different outputs. More on this in Section \ref{sec:chaotic}.} In this case, the augmented inputs consist of $m$ independent blocks that are each based upon the $n$ original inputs (see Figure \ref{fig:linear_vs_random}, right).

\paragraph{ReLU, Leaky ReLU, Absolute value} More usual non-linear activations like ReLU \cite{nair2010rectified, glorot2011deep}, leaky ReLU \cite{maas2013rectifier} (and its variants \cite{xu2015empirical}) or absolute value lead to intermediate $\tilde{\Sigma}$'s, with extra-diagonal matrices that are weaker than $\Sigma$ but stronger than 0. If $\eta$ can be expressed as:

\begin{equation}
\eta(z) = \alpha \mathbbm{1}_{z\leq0} + \beta \mathbbm{1}_{z>0},
 \end{equation}
  
\noindent where $\alpha, \beta$ are normalized such that $\alpha^2 + \beta^2 = 2$, then one can show that if $z_i$ and $z_j$ are independent random variables that are symmetric around 0, then for any variables $y_k$, $y_l$ one has\footnote{The approximation becomes exact if $y_k, y_l$ are also independent from $z_i, z_j$, which becomes a good approximation for $m \gg 1$.}:

\begin{equation}
\begin{cases}
\mathbb{E}\left[\eta(z_i)^2y_k y_l\right] = \mathbb{E}\left[y_k y_l\right],\\
\mathbb{E}\left[\eta(z_i)\eta(z_j)y_k y_l\right] \simeq\nu \mathbb{E}\left[y_k y_l\right],
\end{cases}
\quad \text{with }\nu = \frac{1}{4}\left(\alpha + \beta \right)^2.
 \end{equation}
 
 \noindent The covariance matrix in the augmented space therefore reads:
 
 \begin{equation}\label{eq:relu}
\tilde{\Sigma} \simeq 
\begin{bmatrix}
&\Sigma &\nu\Sigma &\hdots &\nu\Sigma\\
&\nu\Sigma &\Sigma& \hdots &\nu\Sigma\\
&\vdots&\vdots&\ddots&\vdots\\
&\nu\Sigma &\nu\Sigma& \hdots &\Sigma
  \end{bmatrix} \in \mathbb{R}^{nm\times nm}.
  \end{equation}

\begin{figure}[!t]
\includegraphics[width=0.5\textwidth]{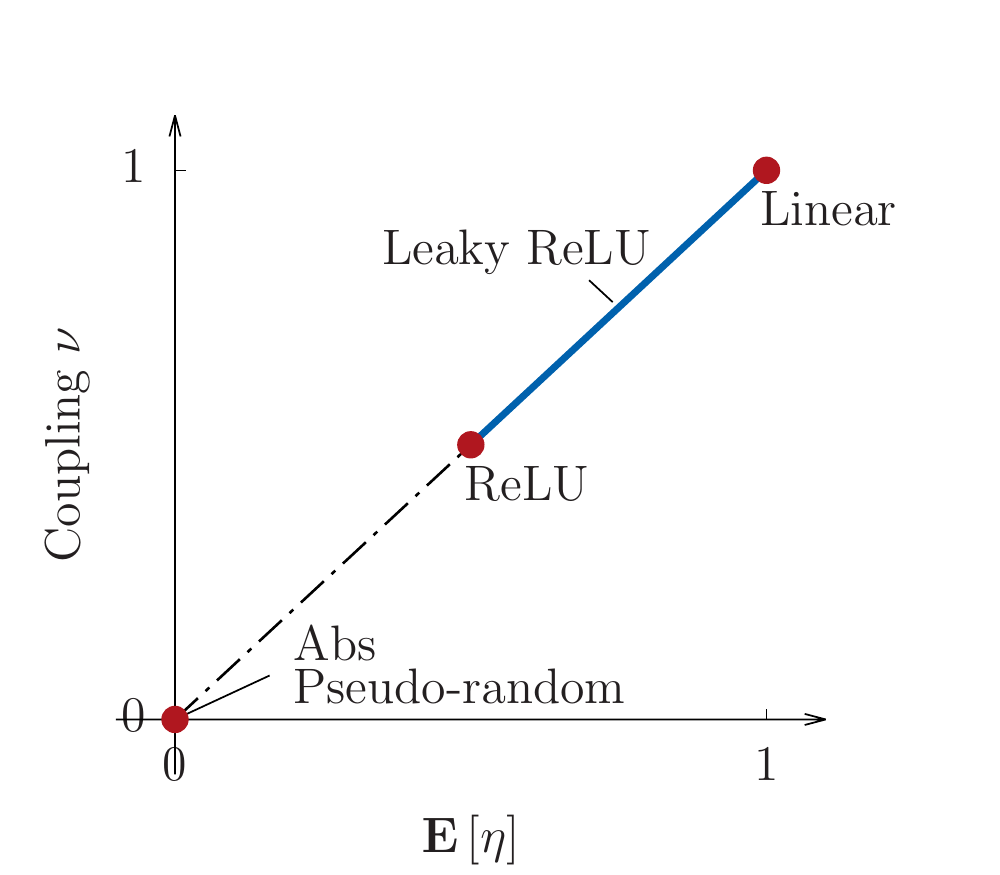}
\centering
\caption{\label{fig:decoupling_graph} The decoupling scale, from fully coupled blocks for the linear layers ($\nu=\mathbb{E}\left[\eta\right]=1$, top right) to fully decoupled blocks for pseudo-random layers ($\nu=\mathbb{E}\left[\eta\right]=0$, bottom left), with Leaky ReLU and ReLU layers in between.}
\end{figure}

\noindent The linear case $\alpha=\beta=1$ yields $\nu=1$ as expected, while a ReLU non-linearity yields $\nu =\frac{1}{2}$ and an absolute value non-linearity with $\alpha=-1$, $\beta=1$ yields $\nu =0$. The leaky ReLU case $\alpha>0$, $\beta >1$ lies somewhere between the linear case and the ReLU case. One could thus be tempted to introduce a decoupling scale to qualify the position of a non-linear activation between the two extreme types discussed above (see also Figure \ref{fig:decoupling_graph}):

\begin{equation}
\underbrace{\text{Linear}}_{\nu=1} < \underbrace{\text{Leaky ReLU}}_{\frac{1}{2}<\nu<1} < \underbrace{\text{ReLU}}_{\nu=\frac{1}{2}} < \underbrace{\text{Abs}}_{\nu=0} < \underbrace{\text{Pseudo-random}}_{\nu=0}
\end{equation}

\noindent Note that here the $\nu$'s are given as an indication and correspond to the case where the $y$'s are independent and symmetric around 0.

\subsection{Linear activations} If $f$ is taken to be linear, which corresponds to the plain form of $\tilde{\Sigma}$ in Eq. (\ref{eq:sigma_plain}) above, it is easy to verify that:

\begin{equation}
\tilde{K} = \frac{1}{\sqrt{m}}\begin{bmatrix} K \\ K \\ ... \\ K \end{bmatrix} 
\end{equation}

\noindent where $K=\Sigma  \frac{\partial PA}{\partial W} \in \mathbb{R}^{n\times p}$ is the capacity matrix that would be obtained by the standard method in the space of the original input $Y$. The capacity allocated to a subspace $\mathcal{S}$ of the input space, which reads $\kappa(\mathcal{S}) = \|K^T S \|_F^2$ in the input space, can then be computed in the augmented input space as:

\begin{equation}
\tilde{\kappa}(\tilde{\mathcal{S}}) = \|\tilde{K}^T \tilde{S} \|_F^2, \quad \text{where } \tilde{S}:= \begin{bmatrix}
&S &0 &\hdots &0\\
&0 &S& \hdots &0\\
&\vdots&\vdots&\ddots&\vdots\\
&0 &0& \hdots &S
  \end{bmatrix}.
\end{equation}

\noindent Indeed, one can then check that this leads to:

\begin{equation}
\tilde{\kappa}(\tilde{\mathcal{S}}) = \kappa(\mathcal{S}).
\end{equation}

\noindent This means that computing the capacity in the augmented input space is equivalent to computing the capacities in the original space -- but with more detailed intermediate results. This can therefore be considered as a generalization of the standard method, which we will use below to compute capacities for non-linear activations.

\subsection{Pseudo-random activations} \label{sec:chaotic}

Tractable expressions for the capacity can also be obtained at the other end of the spectrum, for what we called pseudo-random activations above. As seen above, this case corresponds to choosing $\eta$ as a realization of some i.i.d. process such that:

\begin{equation}
\mathbb{E}(\eta_z)=0, \quad \mathbb{E}(\eta_{z_1}\eta_{z_2}) = \delta_{z_1-z_2}\quad \text{for $z$, $z_1$, $z_2 \in \mathbb{R}$}.  
\end{equation}

\noindent  Let us recall the expression of the network output:

\begin{equation}
\varphi(Y) = A^T f(P^TY) = A^T\tilde{P}^T\tilde{Y}.
\end{equation}






  
\noindent where only the last layer $A$ is learned. 
If we decompose the Gram matrix as $\frac{\partial A}{\partial W}\frac{\partial A^T}{\partial W} = Q^\phi\Lambda Q^{\phi T}$ and note $K^\phi$ the columns of $Q^\phi$ that correspond to the non-zero eigenvalues, then the optimality criterion on the loss $\mathcal{L}$ can be written as:
 
   \begin{equation}
K^{\phi T} \tilde{P}^T \tilde{\Sigma}  (\tilde{P}A - \tilde{P}A^*)= 0 \quad \Leftrightarrow\quad \left(\tilde{\Sigma} \tilde{P} K^\phi\right)^T  .\tilde{X} = 0.
\end{equation}
  
\noindent The matrix $\tilde{\Sigma} \tilde{P} K^\phi $ does not have orthonormal columns in general, and therefore a further orthonormalization is necessary to obtain the expression of the capacity. However, in the specific case where $\Sigma = \sigma^2 \mathbb{I}$ (i.e. where the layer inputs are i.i.d.) this simplifies as:

   \begin{equation}
 \left(\tilde{P} K^\phi\right)^T  .\tilde{X}= 0,
\end{equation}
  
\noindent  and the orthonormality of the columns of $\tilde{P}K^\phi$ stems from that of $\tilde{P}$ (which we have assumed in Section \ref{sec:single_layer}) and $K^\phi$. We can then write the expression for the capacity in the space of the input $\tilde{Y}$ as:

$$\tilde{\kappa}(\tilde{\mathcal{S}}) = \|\tilde{K}^T \tilde{S} \|_F^2, \quad \tilde{K} := \tilde{P}K^\phi.$$

\noindent One can finally define the capacities along the natural dimensions of the input space $\kappa_1, ..., \kappa_m$:

\begin{equation}
\kappa_i = \|K^{\phi T} \tilde{P}^T \tilde{S}_i \|_F^2, \quad \tilde{S}_i := 
\begin{bmatrix}
&e_i &0 &\hdots &0\\
&0 &e_i& \hdots &0\\
&\vdots&\vdots&\ddots&\vdots\\
&0 &0& \hdots &e_i
  \end{bmatrix} \in\mathbb{R}^{nm\times m},
 \end{equation}
 
\noindent  which aggregates the capacities across all $\eta_j y_i$ for a given $i$, and leads to:

\begin{equation}\label{eq: capacity_propagation}
\kappa_i = \sum_{j=1}^m \|K^{\phi T} p_{ij} e_j \|_F^2 =  \sum_{j=1}^m p_{ij}^2 \kappa^\phi_j,
 \end{equation}

\noindent where the last equality comes from the expression of the capacity in the feature space $\kappa^\phi(\mathcal{S}^\phi) = \|\left(K^\phi\right)^T S^\phi \|_F^2$. The above expression can be rewritten as:

\begin{equation}\label{eq:capacity_layer_operator}
\kappa = D.\kappa^{\phi}
\end{equation}

\noindent where $D := P \circ P \in \mathbb{R}^{n\times m}$ is a column-stochastic matrix which preserves the total capacity, and we have defined the spatial capacities along the canonical dimensions of the input and feature spaces, $\kappa = \left(\kappa_1, ..., \kappa_n\right)$ and $\kappa^\phi = (\kappa^\phi_1, ..., \kappa^\phi_m)$. According to this formula, the capacity $\kappa^\phi_j$ allocated to a given component in the feature space propagates to its associated inputs, according to the corresponding (normalized) square coefficient of $P$.\footnote{The case where the columns of $P$ are not normalized is equivalent, but requires to replace $p_{ij} \to \frac{p_{ij}}{\sqrt{\sum_{j=1}^n p_{ij}^2}}$.} Note that this transformation conserves the total capacity: for $\mathcal{S}=\mathbb{R}^n$, one has $\kappa(\mathbb{R}^n) = \sum_{i=1}^n \kappa_i =\sum_{i=1}^m \kappa_i^\phi  = \kappa^\phi(\mathbb{R}^m)$. 

The fact that capacity allocation to a layer inputs only depends on the capacity allocation to the layer outputs is highly interesting. This means that the equation that describes capacity propagation through layers might hold more universally than in the above case where it was derived (namely a $L^2$ loss function and a linear last layer): rather, it might well be valid for any feature space capacity $\kappa^\phi_j$, however derived. We leave this question open for future work.

\section{Multiple neural network layers} \label{sec:multiple_layer}

The capacity propagation equation (\ref{eq:capacity_layer_operator}) obtained in the single layer case can be easily generalized in the multi-layer case, as we now show.

\subsection{General formula} One can extend the above analysis in the case where $\phi$ is a composition of layers:

\begin{equation}
\phi(Y) = \phi_L \circ ... \circ \phi_1(Y), \quad \text{where}\quad \forall l, ~ \phi_l(Y) = f_l(P_l^TY),
\end{equation}

\noindent where $f_l(\cdot)$ is a pointwise non-linear function and $P_l\in\mathbb{R}^{n_{l-1}\times n_l}$ is a linear transformation such that $p_i$ are all distinct and normalized to $\|p_i \|_2 = 1$, as above. Compared with the previous section, we now have $n_L \leftrightarrow m$ and $L\leftrightarrow \phi$ at the uppermost layer. From now on, we will focus on the pseudo-random case unless mentioned otherwise. From Eq. (\ref{eq:capacity_layer_operator}), one can write the spatial capacities at layer $l-1$ as a function of the spatial capacities\footnote{This holds for pseudo-random activations if the network inputs are i.i.d., as in that case the i.i.d. behaviour propagates to all layers.} at layer $l$:



\begin{equation}\label{eq:capacity_layer_operator_single}
\kappa^{l-1} = D_l.\kappa^{l}
\end{equation}

\noindent where $D_l := P_l \circ P_l \in \mathbb{R}^{n_{l-1}\times n_l}$. 
By recursion, one finds:

\begin{equation}\label{eq:capacity_layers_operator_multiple}
\kappa^{0} = D_1 ... D_L\kappa^L
\end{equation}

\begin{figure}[!t]
\def\layersep{0.8cm}
\def\numnodes{15}
\def\middlenode{8}
\def\nodesep{0.5}

\def\ipos{6}
\def \offset{\ipos*\nodesep}

\newcommand\mydots{\makebox[3em][c]{.\hfil.\hfil.}}

\begin{tikzpicture}[shorten >=1pt,->,draw=black!60, node distance=\layersep]
    \tikzstyle{every pin edge}=[<-,shorten <=1pt]
    \tikzstyle{neuron}=[circle,draw=black!80,minimum size=10pt,inner sep=0pt,line width=0.2mm]
    \tikzstyle{input neuron}=[neuron,diagonal fill={blue!0}{yellow!50}];
    \tikzstyle{invisible neuron}=[neuron,minimum size=0pt];
    \tikzstyle{output neuron}=[neuron,fill=red!60];
    \tikzstyle{hidden neuron}=[neuron];
    \tikzstyle{annot} = [ text centered]

    \foreach \name / \y in {1,...,\numnodes}
        \node[input neuron] (I-\name) at (0,\offset-\nodesep*\y) {};
  
        \node[annot, left of=I-1] (annot-1) at (0,\offset-\nodesep*1) {\#1};
        \node[annot, left of=I-2] (annot-2) at (0,\offset-\nodesep*2) {\#2};
        \node[annot, left of=I-\ipos] (annot-i) at (0,\offset-\nodesep*\ipos) {\#i};
        \node[annot, left of=I-\numnodes] (annot-N) at (0,\offset-\nodesep*\numnodes) {\#N};
            \path[-] (annot-2) -- node[auto=false]{\vdots} (annot-i);
            \path[-] (annot-i) -- node[auto=false]{\vdots} (annot-N);

    \foreach \name / \y in {1,...,\numnodes}
        \node[hidden neuron] (H1-\name) at (\layersep,\offset-\nodesep*\y) {};

    \foreach \name / \y in {1,...,\numnodes}
        \node[invisible neuron] (Z1-\name) at (2*\layersep,\offset-\nodesep*\y) {};
        
    \foreach \name / \y in {1,...,\numnodes}
        \node[invisible neuron] (Z2A-\name) at (5*\layersep,\offset-\nodesep*\y) {};
        
    \foreach \name / \y in {1,...,\numnodes}
        \node[hidden neuron] (H2-\name) at (6*\layersep,\offset-\nodesep*\y) {};

    \foreach \name / \y in {1,...,\numnodes}
        \node[invisible neuron] (Z2B-\name) at (7*\layersep,\offset-\nodesep*\y) {};

    \foreach \name / \y in {1,...,\numnodes}
        \node[invisible neuron] (Z3A-\name) at (10*\layersep,\offset-\nodesep*\y) {};
        
    \foreach \name / \y in {1,...,\numnodes}
        \node[hidden neuron] (H3-\name) at (11*\layersep,\offset-\nodesep*\y) {};

    \foreach \name / \y in {1,...,\numnodes}
        \node[invisible neuron] (Z3B-\name) at (12*\layersep,\offset-\nodesep*\y) {};

    \foreach \name / \y in {1,...,\numnodes}
        \node[invisible neuron] (Z4A-\name) at (15*\layersep,\offset-\nodesep*\y) {};
        
    \foreach \name / \y in {1,...,\numnodes}
        \node[hidden neuron] (H4-\name) at (16*\layersep,\offset-\nodesep*\y) {};
               
    \node[output neuron,right of=H4-\middlenode, node distance=1cm] (O) {};
    
 
  \node[hidden neuron, diagonal fill={blue!50}{yellow!50}] (Ifill-\ipos) at (0*\layersep,\offset-\nodesep*\ipos) {};
  \foreach \pos in {\ipos-\nodesep, \ipos+\nodesep}
  \node[hidden neuron, diagonal fill={blue!20}{yellow!50}] (Ifill-\pos) at (0*\layersep,\offset-\nodesep*\pos) {};
  \foreach \pos in {\ipos-2*\nodesep, \ipos+2*\nodesep}
  \node[hidden neuron, diagonal fill={blue!8}{yellow!50}] (Ifill-\pos) at (0*\layersep,\offset-\nodesep*\pos) {};
  \foreach \pos in {\ipos-3*\nodesep, \ipos+3*\nodesep}
  \node[hidden neuron, diagonal fill={blue!4}{yellow!50}] (Ifill-\pos) at (0*\layersep,\offset-\nodesep*\pos) {};
  \foreach \pos in {\ipos-4*\nodesep, \ipos+4*\nodesep}
  \node[hidden neuron, diagonal fill={blue!2}{yellow!50}] (Ifill-\pos) at (0*\layersep,\offset-\nodesep*\pos) {};

  \node[hidden neuron, fill=blue!50] (H1fill-\ipos) at (1*\layersep,\offset-\nodesep*\ipos) {};
  \foreach \pos in {\ipos-\nodesep, \ipos+\nodesep}
  \node[hidden neuron, fill=blue!20] (H1fill-\pos) at (1*\layersep,\offset-\nodesep*\pos) {};
  \foreach \pos in {\ipos-2*\nodesep, \ipos+2*\nodesep}
  \node[hidden neuron, fill=blue!8] (H1fill-\pos) at (1*\layersep,\offset-\nodesep*\pos) {};
  \foreach \pos in {\ipos-3*\nodesep, \ipos+3*\nodesep}
  \node[hidden neuron, fill=blue!4] (H1fill-\pos) at (1*\layersep,\offset-\nodesep*\pos) {};
  \foreach \pos in {\ipos-4*\nodesep, \ipos+4*\nodesep}
  \node[hidden neuron, fill=blue!2] (H1fill-\pos) at (1*\layersep,\offset-\nodesep*\pos) {};

  \node[hidden neuron, fill=blue!60] (H2fill-\ipos) at (6*\layersep,\offset-\nodesep*\ipos) {};
  \foreach \pos in {\ipos-\nodesep, \ipos+\nodesep}
  \node[hidden neuron, fill=blue!25] (H2fill-\pos) at (6*\layersep,\offset-\nodesep*\pos) {};
  \foreach \pos in {\ipos-2*\nodesep, \ipos+2*\nodesep}
  \node[hidden neuron, fill=blue!10] (H2fill-\pos) at (6*\layersep,\offset-\nodesep*\pos) {};
  \foreach \pos in {\ipos-3*\nodesep, \ipos+3*\nodesep}
  \node[hidden neuron, fill=blue!5] (H2fill-\pos) at (6*\layersep,\offset-\nodesep*\pos) {};

  \node[hidden neuron, fill=blue!90] (H3fill-\ipos) at (11*\layersep,\offset-\nodesep*\ipos) {};
  \foreach \pos in {\ipos-\nodesep, \ipos+\nodesep}
  \node[hidden neuron, fill=blue!40] (H3fill-\pos) at (11*\layersep,\offset-\nodesep*\pos) {};
  \foreach \pos in {\ipos-2*\nodesep, \ipos+2*\nodesep}
  \node[hidden neuron, fill=blue!12] (H3fill-\pos) at (11*\layersep,\offset-\nodesep*\pos) {};
   
  \node[hidden neuron, fill=blue] (H4fill-\ipos) at (16*\layersep,\offset-\nodesep*\ipos) {};


    \foreach \source in {1,...,\numnodes}
            \path[-] (I-\source) edge (H1-\source);
    \foreach \source[evaluate=\source as \dest using int(\source+1)] in {1,...,14}
            \path[-] (I-\source) edge (H1-\dest);
    \foreach \source[evaluate=\source as \dest using int(\source-1)] in {2,...,\numnodes}
            \path[-] (I-\source) edge (H1-\dest);

    \foreach \source in {1,...,\numnodes}
            \path[-] (H1-\source) edge (Z1-\source);
    \foreach \source[evaluate=\source as \dest using int(\source+1)] in {1,...,14}
            \path[-] (H1-\source) edge (Z1-\dest);
    \foreach \source[evaluate=\source as \dest using int(\source-1)] in {2,...,\numnodes}
            \path[-] (H1-\source) edge (Z1-\dest);

            \path[-] (Z1-\middlenode) -- node[auto=false]{\mydots} (Z2A-\middlenode);
            \path[-] (Z1-4) -- node[auto=false]{\mydots} (Z2A-4);
            \path[-] (Z1-12) -- node[auto=false]{\mydots} (Z2A-12);

    \foreach \source in {1,...,\numnodes}
            \path[-] (Z2A-\source) edge (H2-\source);
    \foreach \source[evaluate=\source as \dest using int(\source+1)] in {1,...,14}
            \path[-] (Z2A-\source) edge (H2-\dest);
    \foreach \source[evaluate=\source as \dest using int(\source-1)] in {2,...,\numnodes}
            \path[-] (Z2A-\source) edge (H2-\dest);
            
    \foreach \source in {1,...,\numnodes}
            \path[-] (H2-\source) edge (Z2B-\source);
    \foreach \source[evaluate=\source as \dest using int(\source+1)] in {1,...,14}
            \path[-] (H2-\source) edge (Z2B-\dest);
    \foreach \source[evaluate=\source as \dest using int(\source-1)] in {2,...,\numnodes}
            \path[-] (H2-\source) edge (Z2B-\dest);

            \path[-] (Z2B-\middlenode) -- node[auto=false]{\mydots} (Z3A-\middlenode);
            \path[-] (Z2B-4) -- node[auto=false]{\mydots} (Z3A-4);
            \path[-] (Z2B-12) -- node[auto=false]{\mydots} (Z3A-12);
            
    \foreach \source in {1,...,\numnodes}
            \path[-] (Z3A-\source) edge (H3-\source);
    \foreach \source[evaluate=\source as \dest using int(\source+1)] in {1,...,14}
            \path[-] (Z3A-\source) edge (H3-\dest);
    \foreach \source[evaluate=\source as \dest using int(\source-1)] in {2,...,\numnodes}
            \path[-] (Z3A-\source) edge (H3-\dest);
                        
    \foreach \source in {1,...,\numnodes}
            \path[-] (H3-\source) edge (Z3B-\source);
    \foreach \source[evaluate=\source as \dest using int(\source+1)] in {1,...,14}
            \path[-] (H3-\source) edge (Z3B-\dest);
    \foreach \source[evaluate=\source as \dest using int(\source-1)] in {2,...,\numnodes}
            \path[-] (H3-\source) edge (Z3B-\dest);
             
            \path[-] (Z3B-\middlenode) -- node[auto=false]{\mydots} (Z4A-\middlenode);
            \path[-] (Z3B-4) -- node[auto=false]{\mydots} (Z4A-4);
            \path[-] (Z3B-12) -- node[auto=false]{\mydots} (Z4A-12);
            
    \foreach \source in {1,...,\numnodes}
            \path[-] (Z4A-\source) edge (H4-\source);
    \foreach \source[evaluate=\source as \dest using int(\source+1)] in {1,...,14}
            \path[-] (Z4A-\source) edge (H4-\dest);
    \foreach \source[evaluate=\source as \dest using int(\source-1)] in {2,...,\numnodes}
            \path[-] (Z4A-\source) edge (H4-\dest);
                                             
    \foreach \source in {1,...,\numnodes}
        \path[-] (H4-\source.east) edge (O.west);
        
        \path[-, draw=blue, line width=0.4mm] (H4-\ipos.east) edge (O.west);

    \node[annot,above of=H2-1, node distance=1cm] (hlk) {Layer $k$};
   \node[annot,above of=H3-1, node distance=1cm] (hll) {Layer $l$};
   \node[annot,above of=H4-1, node distance=1cm] (hll) {Last layer $L$};
    \node[annot,above of=I-1, node distance=1cm] {Inputs};
    \node[annot,right of=O, node distance=1cm] {Output};

   
   \path[line width=0.3mm, -{Latex[length=3mm,width=3mm]}] (16*\layersep, \offset - \numnodes*\nodesep - 2*\nodesep) edge (0,  \offset - \numnodes*\nodesep - 2*\nodesep) ;
    
    \node[annot] at (8*\layersep, \offset - \numnodes*\nodesep - 3*\nodesep) (legend) {Capacity propagation through layers} ;
    

    \foreach \i in {20, ..., 16}{
 		 \clip plot [smooth, tension=1.4] coordinates { (0,0.3*\i*\nodesep) (16*\layersep, 0) (0,-0.3*\i*\nodesep) };
 		\clip  (0,\offset-\nodesep*\numnodes-\nodesep) rectangle (16*\layersep,\offset-\nodesep) ;
  		\path [fill=blue!30, opacity=0.02] (0,-0.3*\i*\nodesep) rectangle (16*\layersep,0.3*\i*\nodesep) ;	}

    \foreach \i in {15, ..., 11}{
 		 \clip plot [smooth, tension=1.4] coordinates { (0,0.3*\i*\nodesep) (16*\layersep, 0) (0,-0.3*\i*\nodesep) };
 		\clip  (0,\offset-\nodesep*\numnodes-\nodesep) rectangle (16*\layersep,\offset-\nodesep) ;
  		\path [fill=blue!30, opacity=0.03] (0,-0.3*\i*\nodesep) rectangle (16*\layersep,0.3*\i*\nodesep) ;	}

    \foreach \i in {20, ..., 11}{
 		 \clip plot [smooth, tension=1.4] coordinates { (0,0.15*\i*\nodesep) (16*\layersep, 0) (0,-0.15*\i*\nodesep) };
 		\clip  (0,\offset-\nodesep*\numnodes-\nodesep) rectangle (16*\layersep,\offset-\nodesep) ;
  		\path [fill=blue!30, opacity=0.03] (0,-0.15*\i*\nodesep) rectangle (16*\layersep,0.15*\i*\nodesep) ;	}
		
    \foreach \i in {10, ..., 1}{
 		 \clip plot [smooth, tension=1.4] coordinates { (0,0.15*\i*\nodesep) (16*\layersep, 0) (0,-0.15*\i*\nodesep) };
 		\clip  (0,\offset-\nodesep*\numnodes-\nodesep) rectangle (16*\layersep,\offset-\nodesep) ;
  		\path [fill=blue!30, opacity=0.03] (0,-0.15*\i*\nodesep) rectangle (16*\layersep,0.15*\i*\nodesep) ;	}
		
   
\end{tikzpicture}

\centering
\caption{\label{fig:schema} Illustration of the backward propagation of capacity through multiple layers. The blue shades indicate how the capacity allocated to the $i$-th node of the last layer (corresponding to a dirac delta $\kappa^L$) is propagated all the way through the input layer, resulting in a diffused $\kappa^0$. Darker shades corresponds to higher capacity per input. In the deep limit, the propagation is governed by a diffusion PDE, with an effective receptive field that grows as the square root of the number of layers (see Section \ref{sec:receptive}).}

\end{figure}
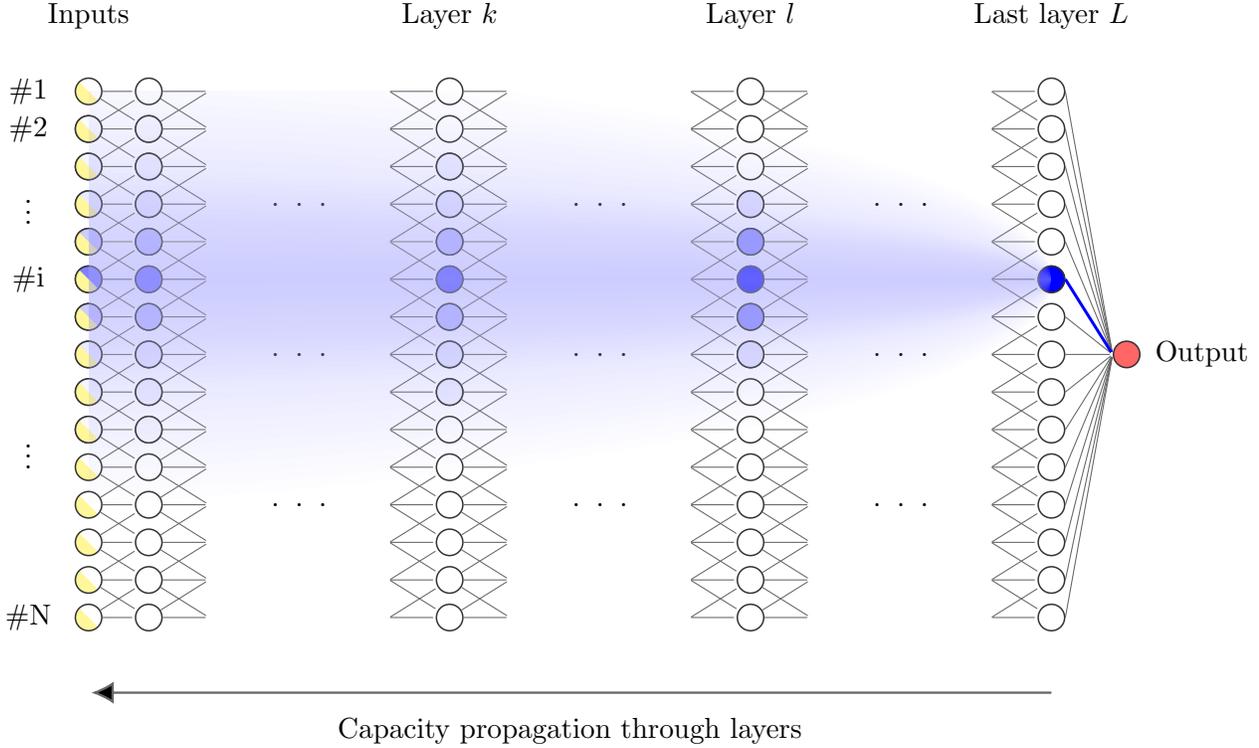

\noindent As mentioned above, the column-stochasticity of the matrices $D_l$ results in the conservation of the total capacity, throughout the intermediate feature spaces in the network -- and all the way to the input space (see Figure \ref{fig:schema} for an illustration).


The remarkable feature of Eq. (\ref{eq:capacity_layers_operator_multiple}) is that it effectively \emph{linearizes the neural network}, with the main difference that this network is applied backwards (in layers) to the capacities rather than forwards to the inputs. We call this network the \emph{capacity network}. Although the results have only been derived under specific hypotheses (essentially, an i.i.d. assumption on the network inputs), this opens the door to analyzing neural networks as linear objects (with the network depth playing the role of time in usual systems), with all the power of linear techniques and a breadth of potential insights. In Section \ref{sec:applications}, we will consider a few examples where studying capacity allows to recover some well-known and empirical results about deep neural networks -- an encouraging start.

\subsection{Limiting behaviours}\label{sec:limiting}

A powerful method for analyzing Eq. (\ref{eq:capacity_layers_operator_multiple}) is to consider the deep limit where:

\begin{equation}
  \begin{cases} 
 \forall i, n_i = n = m ,\\
  L = 1 / \epsilon,\\
 D_1 = ... = D_L = \mathbb{I} + \epsilon \Delta,
   \end{cases} 
\end{equation}
 
 \noindent where $\Delta$ is a matrix of order 1 (assumed to be layer-independant for simplicity) and $\epsilon \to 0$. This amounts to considering deep networks with a large number of layers and a ``residual'' shape for the weights \cite{he2016deep} such that each layer only applies a small transformation to its inputs. If we note $\pi(t) = \kappa^{\frac{1 - t}{\epsilon}}$, such that $\pi(0) = \kappa^L$ and $\pi(1) = \kappa^0$, then $\pi$ satisfies the Kolmogorov forward equation \cite{gardiner2009stochastic}:

\begin{equation}\label{eq:capacity_ODE}
\pi'(t) = \Delta \pi(t).
\end{equation}



\noindent If $\Delta$ is a 1-dimensional local, translation-invariant operator that is non-zero only for $j\in \{i-1, i, i+1\}$, then it can be approximated by the following PDE:

\begin{equation}\label{eq:capacity_PDE}
\pi'(t) = - v\frac{\partial \pi}{\partial x} + \mathcal{D}\frac{\partial^2 \pi}{\partial x ^2}.
\end{equation}

\noindent The solution to this PDE can be written as:

\begin{equation}\label{eq:capacity_PDE_solution}
\pi(t, x) =\frac{1}{\sqrt{4\pi \mathcal{D}t}} \int_{-\infty}^\infty  e^{-\frac{(x - y - vt)^2}{4\mathcal{D}t}} \pi(0, y) \mathrm{d}y
\end{equation}

\noindent If the $D_l$'s are not all equal but random, by symmetry one should obtain $v=0$ (cf \cite{luo2016understanding}), and therefore we obtain in the space of capacities:


\begin{equation}\label{eq:capacity_final_solution}
\kappa^l(x) =\frac{1}{\sqrt{4\pi \mathcal{D}t}} \int_{-\infty}^\infty  e^{-\frac{(x-y)^2}{4\mathcal{D}t}} \kappa^L(y) \mathrm{d}y,
\end{equation}

\noindent where $t = (L - l) / L$.


\subsection{Differential layers} 

Another case that yields the above capacity propagation is the following variant of residual layers, where the non-linear activation is applied to the \emph{residual} only:

\begin{equation}\label{eq:diff_layer}
\phi_l(Y) = Y + \sqrt{\epsilon} f_l(P_l^TY)
\end{equation}

\noindent We call this a differential layer, as it learns exactly the difference between its inputs and its outputs. In this case, the neural network itself can be written as a differential equation in the deep limit. If we note $\epsilon = 1 / L$, $t = l / L$ where $l$ is the layer number and $y(t) :=  \phi_l \circ ... \circ \phi_1(Y)$, one can rewrite Eq. (\ref{eq:diff_layer}) as:

\begin{equation}
y(t + \epsilon) - y(t)= \sqrt{\epsilon} f_l(P^T_{t/\epsilon} y),
\end{equation}

\noindent which is akin to a non-linear \emph{forward} differential equation applied to the input in the $\epsilon \to 0$ limit (remember \textit{a contrario} that the capacity PDE is \emph{backward} in layers).  If $P$ is local as in the previous section, then this can be approximated by the following equation in the 1-dimensional case:

\begin{equation}\label{eq:network_PDE}
y(t + \epsilon) - y(t) =  \sqrt{\epsilon} f\left(\nu^Y_t y - v^Y_t\frac{\partial y}{\partial x} + \mathcal{D}^Y_t\frac{\partial^2 y}{\partial x ^2}\right).
\end{equation}

%
%

\noindent The corresponding capacity PDE in the pseudo-random case is derived in Appendix \ref{sec:diff_pde}, and reads:

\begin{equation}\label{eq:capacity_PDE_diff}
\pi'(t) =2 [\mathrm{d}\mathcal{D}^Y_t, \mathrm{d}\mathcal{D}^Y_t] \frac{\partial^2 \pi}{\partial x ^2},
\end{equation}

\noindent which is similar to the capacity propagation PDE obtained for the type of residual layers studied in the previous section (Eq. \ref{eq:capacity_PDE}). The nice additional property in the differential case is that the diffusion coefficient can be linked explicitly to the diffusion coefficient of the SDE that the forward neural network itself implements (Eq. \ref{eq:network_PDE}).




\section{Applications to theoretical analysis of neural networks}\label{sec:applications}

Eqs. (\ref{eq:capacity_layers_operator_multiple}) and (\ref{eq:capacity_final_solution}) for the propagation of capacity allocation through neural network layers open the door to a number of highly interesting analyses, which uncover some established theoretical and empirical facts about (deep) neural networks. In this section we use the capacity framework to study two questions: the size of the effective receptive field of a deep neural network, and the shattering problem \cite{balduzzi2017shattered}.

\subsection{The effective receptive field of deep networks} \label{sec:receptive}

The above results show that the spatial capacity $\kappa^L$ of the uppermost layer is being spatially diffused as it passes through the layers. For example, if the spatial capacity at the uppermost layer $\kappa^L(x)$ takes the form of a Dirac delta $\kappa^L(x) = \kappa^L \delta_{x - x_0}$, then the resulting spatial capacity in intermediate feature spaces is Gaussian:

\begin{equation}\label{eq:capacity_dirac}
\kappa^l(x) =\frac{e^{-\frac{(x-x_0)^2}{4\mathcal{D}t}}}{\sqrt{4\pi \mathcal{D}t}}  ~\kappa^L,
\end{equation}

\noindent where $t = (L - l) / L$. The larger $L-l$, i.e. the number of layers traversed, the larger the diffusion region (which one might think of as the effective receptive field -- see Figure \ref{fig:schema}). The size of the diffusion region scales as the \emph{square root} of the number of layers that are traversed since $L - l~\propto~ t$ (rather than linearly, as one might naively think), and in the amplitude of $\Delta$ (which is $\propto ~\mathcal{D}$). These results are consistent with the ones of \cite{luo2016understanding}, who found such a Gaussian distribution with square root scaling both theoretically in the linear and the random weights cases, and empirically in the general non-linear case. Interestingly, their assumptions and their derivation of the results are very similar to the ones used here -- but in the space of gradients rather than capacities.
Note that we described the 1-dimensional case here, but such diffusion equation with square root spatial scaling can be derived in arbitrary numbers of dimensions. 

\subsection{Residual layers and shattered capacity} Interestingly, the limiting behaviour described above is very reminiscent of ResNets \cite{he2016deep}, where the authors:

\begin{enumerate}[(i)]
\item introduced residual connections such that $D_l = \mathbb{I} + \epsilon \Delta_l $ becomes a good approximation for each layer,
\item use size 3 convolutions, such that $\Delta$ is a local, translation-invariant operator that is non-zero for $j\in \{i-1, i, i+1\}$ only.\footnote{Although their use case was 2-dimensional, which would requires cross-shaped instead of square filters.}
\end{enumerate}

In this setting, a significant local structure remains at $l=0$ (according to Eqs. (\ref{eq:capacity_final_solution}) and  (\ref{eq:capacity_dirac})) as the capacity is not ``shattered'' by passing through the layers. If we decompose the influence of $\kappa^L$ on the intermediate capacity $\kappa^l$ into the different paths:

\begin{equation}\label{eq:capacity_paths}
\kappa^{l}_{i_l} = \sum_{i_L} \kappa^L_{i_L} \sum_{i_{l+1}, ..., i_{L-1}} \underbrace{(D_{l+1})_{i_l i_{l+1}}(D_{l+1})_{i_{l+1}i_{l+2}} ... (D_{L})_{i_{L-1}i_L}}_{\text{Weight } w_{i_l, ..., i_L}},
\end{equation}

\noindent then the path with the highest influence is for $i_l = i_{l+1} = ... = i_{L} (:=i)$ and its weight is 

\begin{equation}
w_\mathrm{max}^i = \prod_{l'=l+1}^{L} (D_{l'})_{ii} = \prod_{l'=l+1}^{L} \left( 1 + \epsilon (\Delta_{l'})_{ii} \right) \simeq e^{\int_0^t \Delta_s^i\mathrm{d} s}, 
\end{equation}

\noindent where we used the simplified notation $\Delta_s^i \leftrightarrow (\Delta_{l})_{ii}$ where $s :=(L-l)/l$. Note that $w_\mathrm{max}^i < 1$ since  $(\Delta_{l'})_{ii} < 0$. Therefore, the weight of the most straightforward path between the last layer capacity $\kappa^L$ and the intermediate layer capacity $\kappa^l$ decays \emph{exponentially} with the number of layers. In the case above where $L$ scales as $1 / \epsilon$,  the matrices $\Delta_l$ are of order 1 and therefore the largest weight $e^{\int_0^t \Delta_s^i\mathrm{d} s}$ is of order 1. 

If, on the other hand, $P$ is uniform with receptive field size $r$ (and thus so is $D$), the weights are uniform and equal to $\left(\frac{1}{r}\right)^L$ -- i.e. they are vanishingly small. This is very reminiscent of the ``shattered gradients'' analysis of \cite{balduzzi2017shattered}, who have shown how ResNets solve the shattered gradients problem by introducing residual connections -- again, using similar methods to ours but applied to the gradients rather than the capacities. 






\section{Conclusion}

In this paper, we have built upon the notion of capacity in the feature space introduced in \cite{donier2018capacity}, and addressed the case where the mapping function is a neural network layer. We have shown that in this case, the capacity in the input space can be computed under some special hypotheses. 
We have derived theoretical results for two limiting ends of the spectrum, namely (i) for linear activations (non-decoupling layers) where the original results are exactly recovered, and (ii) for pseudo-random activations (fully decoupling layers) in which the capacity in the input space can be easily computed as a linear transformation of the capacity in the feature space. 

When extending the scope to deep neural networks, we have shown that the propagation of capacity through the network layers follows a linear Markov equation in the pseudo-random case, which can be turned into a PDE under appropriate scalings. The remarkable aspect of this result is that it allows to analyze a non-linear neural network in terms of some linear counterpart (the \emph{capacity network}) which can then be analyzed using powerful linear methods. Although we have only obtained results in special cases and more work is required to understand the regime where the markovian propagation is valid, the fact that it allowed us to recover some theoretical and empirical findings that have been reported in earlier studies is encouraging -- and suggests that capacities and gradients may be two sides of the same coin in the pseudo-random case. More generally, the present analysis suggests that the notion of capacity may provide a new vantage point for studying theoretically a number of fundamental questions in deep learning, and provide a more principled and quantitative framework for architecture design. 

\section*{Acknowledgements} The author would like to thank Martin Gould, Marc Sarfati and Antoine Tilloy for their very useful comments on the manuscript.

\bibliography{Capacity_allocation_non_linear}{}
\bibliographystyle{plain}

\clearpage

\appendix

\section{Generalizing the results} \label{sec:P}

Two challenges arise when trying to generalize the above results:

\begin{enumerate}[(i)]
\item For general non-linearities, $\eta$ depends on $Y$. Therefore, the transition from the augmented input space to the original input space is trickier: it is not clear that one can still add the capacities as if the space was separable.
\item When computing the capacity allocated by lower layers (e.g. if $P$ also becomes a variable), the $\eta$'s used for later layers depend on the parameter of the layer considered. Therefore $\tilde{\Sigma}$ does too, which must be taken into account in the calculation of $\frac{\partial \mathcal{L}}{\partial W}$ in Eq. \ref{eq:capacity_augmented}.
\end{enumerate}

One possible solution to (ii) is to restrict the scope to locally linear functions $f$, i.e. functions that can be written as:

\begin{equation}
f(y) = \eta_y y, \quad \text{with } f'(y) = \eta_y,
\end{equation}

\noindent where $\eta_y$ is piecewise constant. In that case, for any set of inputs $\{Y^i\}_{i=1,\ldots,n}$, an infinitesimal perturbation of the network parameters will keep all the $\eta_y$'s constant for all $Y_i$ if the perturbation $\epsilon$ is small enough. The possible difficulty is that the more inputs $Y$ are being considered, the smaller $\epsilon$: the result might only be valid in some precise limit where the expectation over $Y$ and the derivative w.r.t. $W$ are being taken jointly -- rather than taking the expectation first, then computing the derivative, as was done in Eq. (\ref{eq:capacity_augmented}). 

Justifying (i) seems somewhat trickier. One possibility is to accept the current formula as some approximation of the true capacity. Hopefully, the dependence of the various $\eta$'s on the inputs $Y$ is much weaker than the dependence of $Y$ on itself. This is more likely to be true for non-linearities like ReLU, leaky ReLU or absolute value, where $\eta$ is a simple binary function of its inputs.

Some results produced when applying Eqs. (\ref{eq:augmented_model}) and (\ref{eq:augmented_model_detail}) to lower layers are consistent with what one would expect for piecewise linear, homogeneous functions (i.e. ReLU / leaky ReLU / Absolute value etc.). For example, we have checked that the formula allows to retrieve the correct total capacity for a few handpicked hierarchical architectures. In these cases, one expects some parameter redundancy across layers as the scales of successive layers compensate each other -- which is exactly what is found. This also suggests that the formulae break down in any other cases (e.g. sigmoid, tanh...), as the equations would find some redundancy when there is in fact none: ReLU-type functions might actually be the only type of functions that allow for the above analysis. This would be consistent with the particular attention that these functions have previously been given in the literature.

\section{Capacity PDE for differential layers} \label{sec:diff_pde}

Equation (\ref{eq:network_PDE}) shows, maybe trivially, that a deep network with differential layers and pseudo-random weights can be seen as a non-linear SDE applied to its inputs. The expression of the capacity is similar to the case described in Section \ref{sec:limiting}, provided one adds variables in the augmented input space corresponding to the un-modified inputs:

\begin{equation}
\varphi(Y) = A^T \left(Y + \sqrt{\epsilon}  f(P^TY)\right) = A^T\tilde{P}^T\tilde{Y},
\end{equation}

\noindent where:

\begin{equation}\label{eq:augmented_model_detail_differential}
\tilde{P} = 
\begin{bmatrix}
&1 &0 &\hdots &0\\
&0 &1 &\hdots &0\\
&0 &0 &\hdots &1\\
&\sqrt{\epsilon}  p_1 &0 &\hdots &0\\
&0 &\sqrt{\epsilon}  p_2& \hdots &0\\
&\vdots&\vdots&\ddots&\vdots\\
&0 &0& \hdots &\sqrt{\epsilon}  p_n
  \end{bmatrix} \in \mathbb{R}^{n(n + 1) \times n}, \quad 
  \tilde{Y} = 
\begin{bmatrix}
y_1\\
 \vdots\\
  y_n\\
\eta_1 y_1\\
 \vdots\\
  \eta_1y_n\\
   \vdots\\
   \eta_ny_1\\  
    \vdots\\
   \eta_ny_n\\  
\end{bmatrix}\in \mathbb{R}^{n(n + 1)}.
\end{equation}

\noindent In the pseudo-random case, this leads to the following expression of the spatial capacity in the input space, under the same hypotheses as in Section \ref{sec:limiting} (essentially, $\Sigma = \sigma^2\mathbb{I}$ and $\|p_i \|_2 = 1$):

\begin{equation}\label{eq: capacity_propagation_differential}
\kappa_i = \frac{1}{1 + \epsilon} \left( \kappa^\phi_i +  \epsilon \sum_{j=1}^m p_{ij}^2 \kappa^\phi_j \right) = \kappa^\phi_i + \frac{\epsilon}{1 + \epsilon} \left( \sum_{j=1}^m p_{ij}^2 \kappa^\phi_j  - \kappa^\phi_i\right)
 \end{equation}

\noindent which gives:

\begin{equation}
\kappa = D \kappa^\phi,\quad D := \mathbb{I} + \frac{\epsilon}{1 + \epsilon}  \Delta \sim \mathbb{I} + \epsilon  \Delta
 \end{equation}

\noindent where $\Delta = P\circ P - \mathbb{I}$. This is very reminiscent of the expressions obtained in Section \ref{sec:limiting}, and can be treated in a similar way to obtain the capacity PDE:

\begin{equation}
\pi'(t) = - v_t\frac{\partial \pi}{\partial x} + \mathcal{D}_t\frac{\partial^2 \pi}{\partial x ^2}.
\end{equation}

\noindent If $P$ fluctuates randomly from layer to layer, then by symmetry one would have $v_t=0$ (as in Section \ref{sec:limiting}) so that the above equation can be rewritten as:

\begin{equation}
\pi'(t) = \mathcal{D}_t\frac{\partial^2 \pi}{\partial x ^2}.
\end{equation}

\noindent From the expression of $\Delta$, one can check that $[\mathrm{d}\mathcal{D}^Y_t, \mathrm{d}\mathcal{D}^Y_t] = \frac{1}{2} \mathcal{D}_t$, where $\mathcal{D}^Y_t$ is the diffusion coefficient of the stochastic process implemented by the forward neural network (see Eq. (\ref{eq:network_PDE})).

\end{document}